# Coach2vec: autoencoding the playing style of soccer coaches

**Paolo Cintia (**University of Pisa)
**Luca Pappalardo** (ISTI-CNR, Pisa)

**Abstract**

Capturing the playing style of professional soccer coaches is a complex, and yet barely explored, task in sports analytics. Nowadays, the availability of digital data describing every relevant spatio-temporal aspect of soccer matches, allows for capturing and analyzing the playing style of players, teams, and coaches in an automatic way. In this paper, we present coach2vec, a workflow to capture the playing style of professional coaches using match event streams and artificial intelligence. Coach2vec extracts ball possessions from each match, clusters them based on their similarity, and reconstructs the typical ball possessions of coaches. Then, it uses an autoencoder, a type of artificial neural network, to obtain a concise representation (encoding) of the playing style of each coach. Our experiments, conducted on soccer-logs describing the last four seasons of the Italian first division, reveal interesting similarities between prominent coaches, paving the road to the simulation of playing styles and the quantitative comparison of professional coaches.

## Introduction

Coaches are often referred to by their playing style as much as their results, and some playing styles have gone down in history, such as "Catenaccio" (H. Herrera and N. Rocco), "Total Football" (J. Cruyff), "Tika-taka" (P. Guardiola), or "Sarriball" (M. Sarri). But, how can we capture a coach's playing style? How can we state how similar two coaches' styles are? These are complex questions that we address in this paper using digital data and artificial intelligence.

Nowadays sensing technologies extract high-fidelity digital data from every match. In particular, soccer-logs describe all the spatio-temporal events that occur during a match (e.g., passes, shots, fouls), opening up for the analysis of different aspects of the "beautiful game" (Pappalardo et al., 2019a). While these data have been largely used for player scouting (Decroos et al., 2019), performance evaluation (Pappalardo et al., 2019b,

Duch et al., 2010), and the style detection of teams (Bialkowski et al., 2014, Cintia et al., 2015, Buldù et al., 2019, Gyarmati and Anguera, 2015, Gyarmati et al., 2014) and players (Decroos and Davis, 2019), there are no mature approaches to capture the style of professional coaches (Hewitt et al., 2016).

This paper presents **coach2vec**, which aims to capture the playing style of professional coaches through a five-step process based on soccer-logs. Coach2vec describes each possession phase in a match with features capturing its duration, speed, position and length. Then, it uses clustering to find groups of similar possession phases, allowing to determine the possession phases that are typical of a specific coach. Coach2vec hence uses an autoencoder, a type of artificial neural network, to learn efficient and concise representation (encoding) of the playing style of coaches in an unsupervised manner.

We apply Coach2Vec to all matches in the last four sessions of the Italian first division, in order to extract profiles for each pair of coaches and teams. The analysis of these profiles reveal interesting similarities between coaches in the Italian first division, a first step towards the simulation of playing styles and the quantitative comparison of professional coaches.

**Methods**

Soccer-logs describe the behavior of two facing teams in a match (Pappalardo et al., 2019). They are collected by annotating manually a set of *events*, i.e., interactions among the players and the ball. Each event is associated with its starting and ending positions, type (pass, shot, etc.) and outcome (e.g., accurate, won, lost, goal).

We design a workflow to detect, from soccer-logs, the behavior of teams. The workflow exploits clustering to aggregate coaches by the similarity of their style and autoencoders to create an embedding of their profiles.

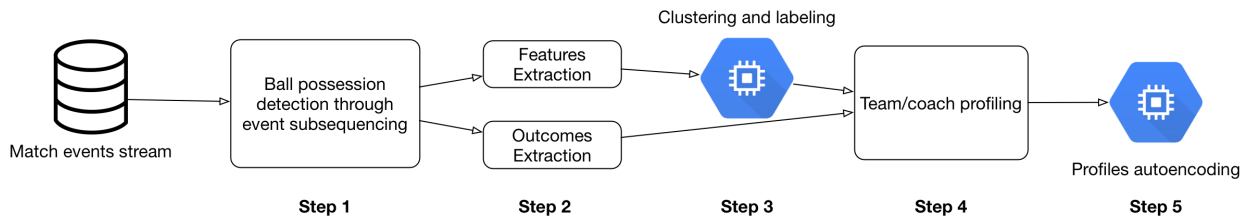

*Figure 1:* **The Coach2Vec workflow.** *Using soccer-logs, we create an encoding of coach profiles using autoencoders and clustering.*

**Step 1.** We extract all the ball possession phases from the matches' soccer-logs. A ball possession phase is a sequence of events related to a portion of the match in which the same team owns the ball.

**Step 2.** For each ball possession, we compute the following features:
- **duration:** difference between the last and the first event;
- **avg_pass_length:** average length of each pass;
- **avg_y:** average value of the y coordinate of the events;
- **start_x:** x coordinate of the first event;
- **speed_step_1/2/3 :** we split each possession phase into three equal time slots and compute the ball speed on the 1st, 2nd or 3rd part of the possession phase.

A possession phase is thus represented as a vector of seven elements, normalized using z-score.

**Step 3.** We use $k$-Means to group the possession phases based on their similarity. We set k=10 based on the elbow method (Figure 2). Figure 3 shows the resulting clusters and the feature values of the centroid. We named the clusters based on the characteristics of their centroid. For example, the centroids with high values for the feature *duration* have been labeled as "long possession": based on speed, we characterize long possessions "long possession final acceleration", when the speed for the 3rd part of the possession is high, "long possession fast approaching" when the speed for the 2nd part is high, or "long possession slow advancing" when speed does not have any particular peak. "Fast bottom build" phases have a low starting $x$ position, and a very high speed in the first third of the possession phase. "Long" ball phases are characterized by the highest average pass length. "High recovery and rebuild" possessions have an high starting $x$ position combined with a very low initial speed: this happens when a team recovers the ball close to the opponent's goal and consolidates the possession by circulating the ball back, instead of directly progressing the ball forward. Possession on right and left flanks are characterized by their average $y$ value. They can be either high recovery or fast building depending on the starting $x$ position

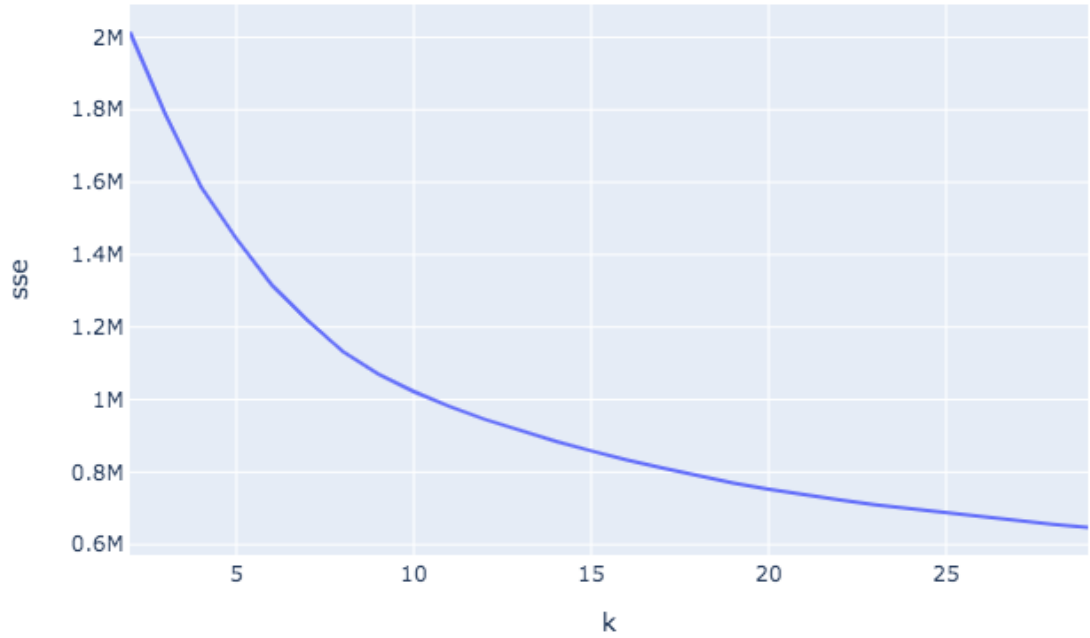

*Figure 2: Sum of Squared Errors (SSE) on varying the number of clusters (k) for ball possession phases clustering*

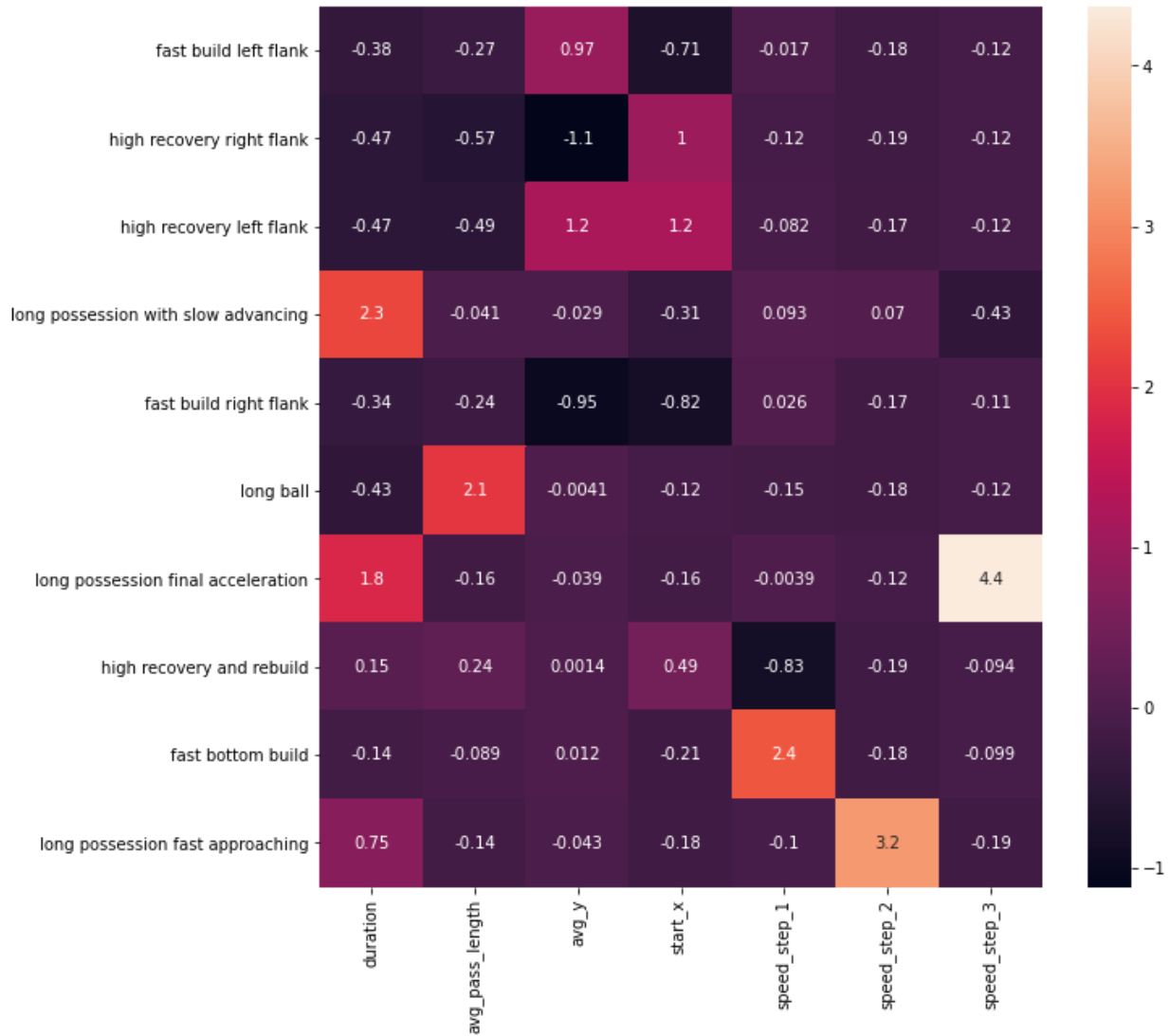

*Figure 3: Feature standardized values for each cluster.*

**Step 4.** We define a team strategy by analyzing the distribution of possession phase types under different contexts. In particular, for all the matches related to a coach managing the same team, we create the following features:

- **Avg ratio per match:** average percentage of possession performed per match, for each possession type;
- **Suffered avg ratio per match:** average percentage of possession conceded to the opponent per match, for each possession type;
- **Avg expected goals per match:** average expected goals per match achieved, for each possession type;

- **Suffered avg expected goals per match:** average expected goals per match achieved, for each possession type;
- **Avg ratio in winning context:** average percentage of possession type performed while the team was leading the game, i.e., when the goal difference is positive in favor of coach's team
- **Avg ratio in drawing context:** average percentage of possession type performed while the team was tying the game, i.e. when the goal difference is zero
- **Avg ratio in losing context:** average percentage of possession type performed while the team was leading the game, i.e. when the goal difference is negative against coach's team

**Step 5.** A coach's profile is then defined as a 7x10 matrix (Figure 4). To look for profile similarities we rely on autoencoders, an unsupervised artificial neural network that learns how to efficiently compress and encode data and reconstruct the data back from the reduced encoded representation to a representation that is as close to the original input as possible.[1] Autoencoders reduce data dimensions by learning how to in the data. Such dimensionality reduction is the key of our coach profile similarity search: by reducing the coach profile to a vector of 5 elements, we can compute similarities among coaches relying on the Euclidean function. Figure 4 shows the architecture of the autoencoder. After a cross validation process, we select "adadelta" as the best optimizer and mean squared error as a loss function. We train the model over 3000 epochs: at the end of training, the loss function converges to a value of 0.01.

---

[1] We also tested classical clustering techniques, finding no valuable results using density-based or distance-based clustering techniques.

| | Fast build left flank | High recovery right flank | High recovery left flank | Long possession slow advancing | Fast build right flank | Long ball | Long possession final acceleration | High recovery and rebuild | Fast bottom build | Long possession fast approaching |
|---|---|---|---|---|---|---|---|---|---|---|
| Avg ratio per match | 0.06 | 0.14 | 0.15 | 0.07 | 0.12 | 0.17 | 0.05 | 0.05 | 0.1 | 0.2 |
| Suffered avg ratio per match | 0.06 | 0.14 | 0.15 | 0.07 | 0.12 | 0.17 | 0.05 | 0.05 | 0.1 | 0.2 |
| Avg expected goals per match | 0.06 | 0.14 | 0.15 | 0.07 | 0.12 | 0.17 | 0.05 | 0.05 | 0.1 | 0.2 |
| Suffered avg expected goals per match | 0.06 | 0.14 | 0.15 | 0.07 | 0.12 | 0.17 | 0.05 | 0.05 | 0.1 | 0.2 |
| Avg ratio in winning context | 0.06 | 0.14 | 0.15 | 0.07 | 0.12 | 0.17 | 0.05 | 0.05 | 0.1 | 0.2 |
| Avg ratio in drawing context | 0.06 | 0.14 | 0.15 | 0.07 | 0.12 | 0.17 | 0.05 | 0.05 | 0.1 | 0.2 |
| Avg ratio in losing context | 0.06 | 0.14 | 0.15 | 0.07 | 0.12 | 0.17 | 0.05 | 0.05 | 0.1 | 0.2 |

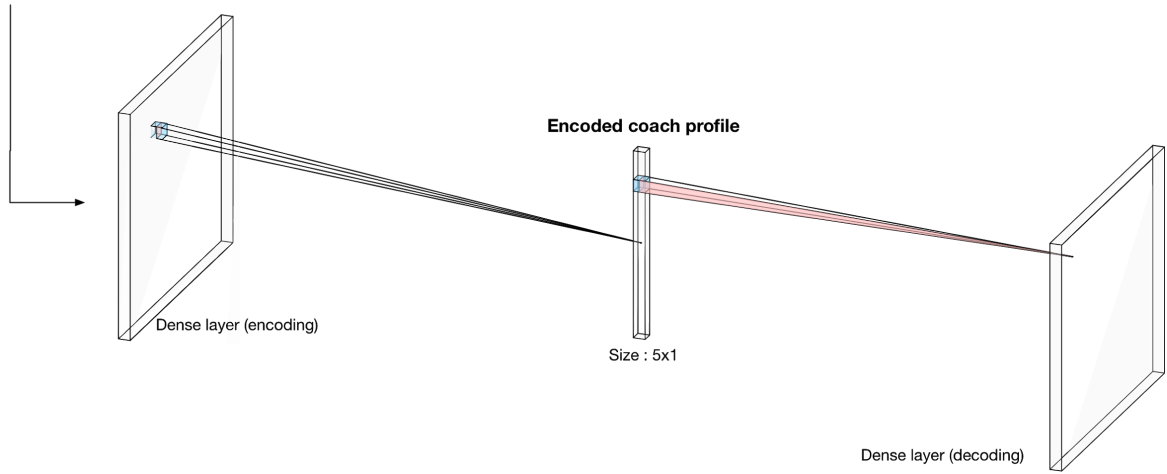

*Figure 4: Autoencoder architecture: encoding and decoding of coach profiles is performed exploiting Dense layers.*

## Results

We apply our workflow on the matches of the last four seasons of the Italian first division. In Figure 5, we show the most similar coaches to Antonio Conte, Carlo Ancelotti, Gianpiero Gasperini, Gennaro Gattuso, Maurizio Sarri and Roberto de Zerbi.

Gasperini (Atalanta) - one of the most celebrated coaches of season 2019/2020 given the striking performance of Atalanta in the national and European competitions - has a pretty unique profile: with respect to the other coaches, the coach with the most similar style (another talented and young coach, De Zerbi) is at a high distance (0.034, see Figure 4).

The well-known style of Maurizio Sarri's Napoli is the most similar to Giampaolo's Sampdoria, another team that had a "Sarriball" like playing style.

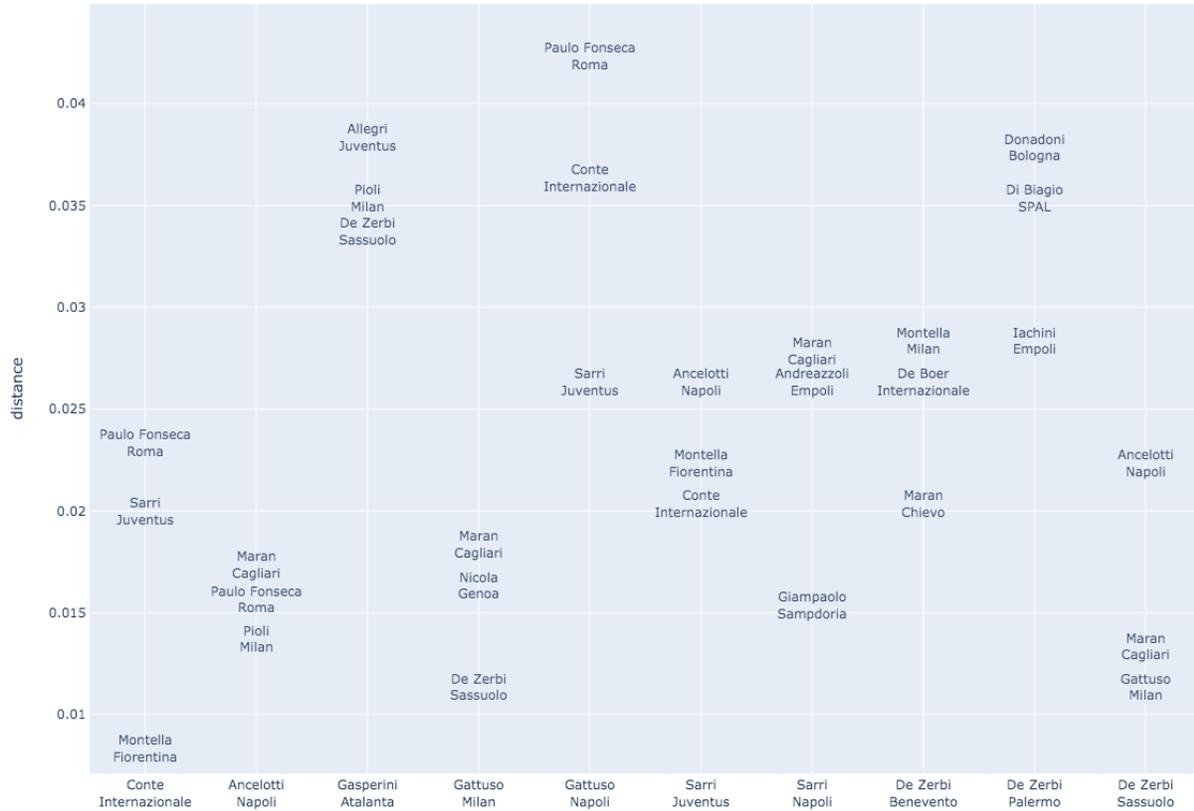

*Figure 5: Coach similarity according to coaches encoded profile distance*

**Practical application**

Coach2Vec is useful to dive into large datasets of soccer-logs and search for specific playing styles, with the possibility of performing queries by similarity. For example, we can find the most similar coach to an input one, or detect the playing style of the last 5 matches of the next opponent, their behavior under different context, and the most similar matches to a reference one. Coach2vec can be used for coach scouting too, a topic that has not been the primary focus of football analytics research. Moreover, our approach is a step toward the speed-up of match analysis: Coach2Vec is capable of detecting changes in a team playing style, with the possibility, for the analyst, to link videos to specific games or part of the games where a specific pattern/style has been detected.